\documentclass[pmlr]{jmlr}% new name PMLR (Proceedings of Machine Learning)

\usepackage{longtable}% for long tables

\usepackage{booktabs}
\usepackage[load-configurations=version-1]{siunitx} % newer version
\theorembodyfont{\upshape}
\theoremheaderfont{\scshape}
\theorempostheader{:}
\theoremsep{\newline}

\jmlrvolume{1}
\jmlryear{2018}
\jmlrworkshop{International Workshop on Cost-Sensitive Learning}

\newcommand{\cNP}[0]{\ensuremath{\mathsf{NP}}\xspace}

\newcommand{\norm}[1]{\left\lVert#1\right\rVert}

\usepackage{todonotes}
\usepackage{enumitem}
\usepackage{multirow}
\renewcommand\vec{\mathbf}
\makeatletter
\renewcommand*\env@matrix[1][*\c@MaxMatrixCols c]{%
  \hskip -\arraycolsep
  \let\@ifnextchar\new@ifnextchar
  \array{#1}}
\makeatother

\graphicspath{{./images/}}

\title[Recognizing Cuneiform Signs Using Graph Based Methods]{Recognizing Cuneiform Signs Using Graph Based Methods}

  \author{\Name{Nils M.\@ Kriege} \Email{nils.kriege@tu-dortmund.de}\\
   \Name{Matthias Fey} \Email{matthias.fey@tu-dortmund.de}\\
   \Name{Denis Fisseler} \Email{denis.fisseler@tu-dortmund.de}\\
   \Name{Petra Mutzel} \Email{petra.mutzel@tu-dortmund.de}\\
   \Name{Frank Weichert} \Email{frank.weichert@tu-dortmund.de}\\
   \addr Department of Computer Science, TU Dortmund, Germany}%

\begin{document}

\maketitle

\begin{abstract}
The cuneiform script constitutes one of the earliest systems of writing and is
realized by wedge-shaped marks on clay tablets. A tremendous number of cuneiform
tablets have already been discovered and are incrementally digitalized and made
available to automated processing.
As reading cuneiform script is still a manual task, we address the real-world application of recognizing cuneiform signs by two graph based methods
with complementary runtime characteristics.
We present a graph model for cuneiform signs together with a tailored distance
measure based on the concept of the graph edit distance.
We propose efficient heuristics for its computation and demonstrate its
effectiveness in classification tasks experimentally.
To this end, the distance measure is used to implement a nearest neighbor
classifier leading to a high computational cost for the prediction phase with
increasing training set size.
In order to overcome this issue, we propose to use CNNs adapted to graphs as an
alternative approach shifting the computational cost to the training phase.
We demonstrate the practicability of both approaches in an experimental
comparison regarding runtime and prediction accuracy.
Although currently available annotated real-world data is still limited, we obtain a high
accuracy using CNNs, in particular, when the training set is enriched by augmented
examples.
\end{abstract}

\section{Introduction}%
\label{sec:intro}

\begin{figure}[tb]
\floatconts%
  {fig:cuneiform}
  {\caption{Example of a cuneiform tablet with tetrahedron shaped markings.}}
  {%
    \null\hfill
    \subfigure[Cuneiform tablet fragment]{\label{fig:cuneiform_tablet}%
       \includegraphics[height=3cm]{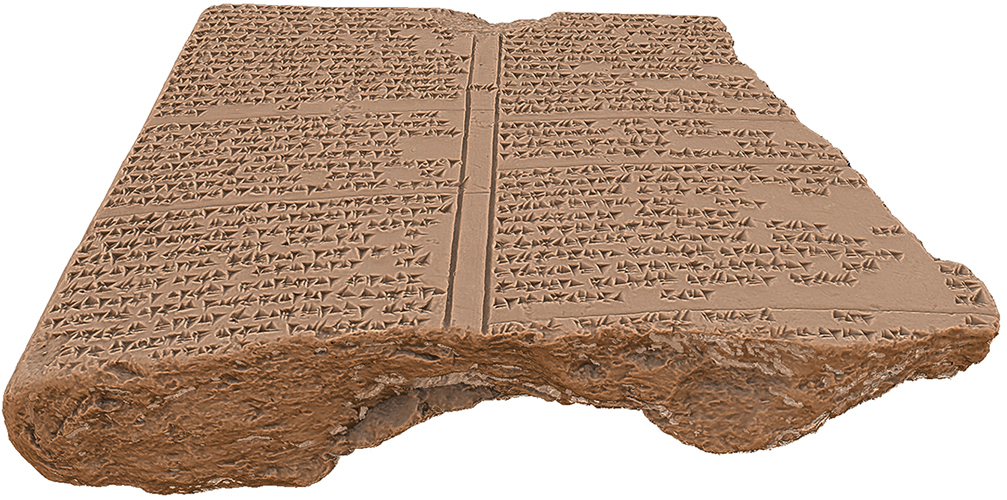}%
    }%
    \hfill
    \subfigure[Tetrahedron shaped markings]{\label{fig:cuneiform_wedge}%
      \includegraphics[height=3cm]{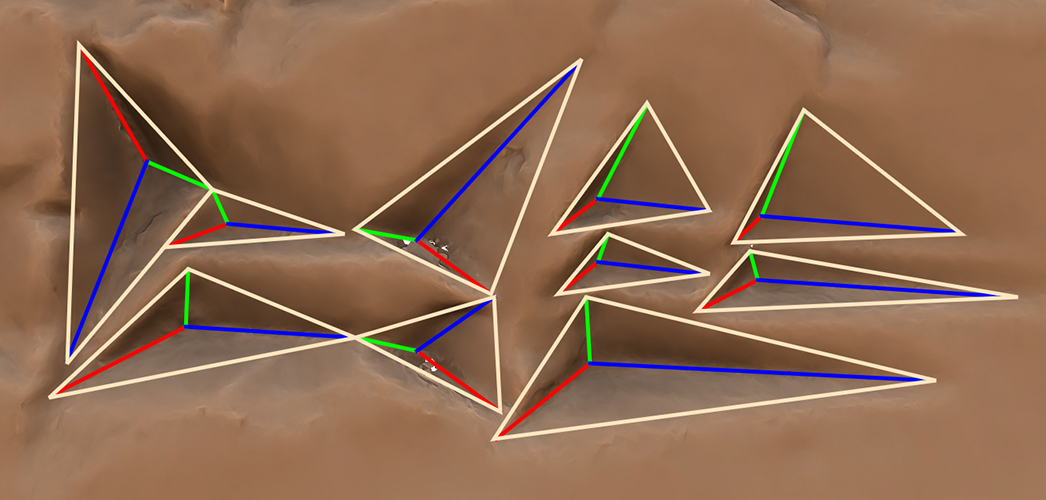}%
    }%
    \hfill\null%
  }
\end{figure}

Besides the Egyptian hieroglyphs, the cuneiform script is the oldest writing system in the world.
Since its emergence at the end of the 4th century BC the cuneiform writing system was in use for more than three millennia by various cultures~\citep{Ruester1989}.
It was eventually replaced by alphabetic writing systems, which were easier to learn.
Cuneiform manuscripts were written predominantly on clay tablets, \emph{cf.} \figureref{fig:cuneiform_tablet}.
These tablets are an important source for revealing early human history, ranging from simple delivery notes over religious texts to contracts between empires.
Cuneiform inscriptions were usually created by impressing an angular stylus into moist clay, resulting in signs consisting of tetrahedron shaped markings, called \emph{wedges} \citep{Fisseler2013}, as shown in~\figureref{fig:cuneiform_wedge}.

Over time, the inventory of cuneiform signs, \emph{i.e.}, groups of wedges, has been classified for various languages with values ranging from well over 1,000 signs for early cuneiform variants.
For Hittite cuneiform, the visual appearance of wedges can be classified depending on the stylus orientation into vertical, horizontal and diagonal, so-called `Winkelhaken' wedge types~\citep{Ruester1989}.
However, depending on the manuscripts state of conservation, it may not be possible to identify a wedge or sign solely based on its geometric appearance, but the geometric context in form of neighboring wedges and signs must also be considered.

Over 500,000 cuneiform fragments have been discovered so far and are digitalized with the help of high-resolution 3D measurement technologies \citep{Rothacker2015}.
The resulting data provides a starting point for automated processing approaches to support philologists in their analysis.
However, many script related aspects of the digitalized data are annotation-bound, as reading and analyzing cuneiform manuscripts is still a manual task, which is difficult and time consuming even for human experts.
Due to the high data acquisition costs, classifiers need to be able to learn from a small number of examples and generalize well to a huge amount of unknown data.

\subsection{Related work}
We briefly summarize the related work on the cuneiform font metrology, the graph edit distance and geometric deep learning.

\paragraph{Cuneiform font metrology}
Automated processing of digital cuneiform data can be conducted on a variety of representations, ranging from photographic reproductions to 3D scanned point clouds and derived representations.
\citet{Fisseler2013} use a fault tolerant wedge extraction based on a tetrahedron model to create large amounts of data for statistical script analysis.
This wedge model includes wedge component based features, like edges and depth points and a heuristic classification of wedge types.
Based on generating 2D contour line vector representations of wedges, \citet{Bogacz2015,Bogacz2015a} proposed two approaches to classify cuneiform signs:
One of them is based on graph representations, which are compared by graph kernels \citep{Bogacz2015a}, and the other computes the optimal assignment between vector representations \citep{Bogacz2015}.
\citet{Rothacker2015} proposed a Bag-of-Features and HMM based approach for automatically retrieving cuneiform structures on image representations from 3D scans.

\paragraph{Graph edit distance}
Graph based methods have a long history in pattern recognition~\citep{Conte2004}.
In this domain, the graph edit distance has been proposed more than 30 years
ago~\citep{Sanfeliu1983}. Since then it has been proven to be a
flexible method for error-tolerant graph matching with applications in numerous
domains~\citep{Conte2004,Stauffer2017}.
\citet{Bunke1997} has shown that computing the graph edit distance generalizes
the classical maximum common subgraph problem, which is well known to be
\cNP-complete.
Due to the high computational costs involved in solving such problems, the graph 
edit distance is usually not determined exactly in practice. 
\citet{Riesen2009a} proposed a method to derive a series of edit
operations from an optimal assignment between the vertices of the two input
graphs. The method can be computed in cubic time, but not necessarily computes
a minimum cost edit path from one graph to the other. However, the results have
been shown to be sufficiently accurate for many applications~\citep{Stauffer2017}.
Recently, a binary linear programming formulation for computing the exact
graph edit distance has been proposed~\citep{Lerouge2017}. The method was shown
to be efficient for several classes of real-world graphs with state-of-the art
general purpose solvers.

\paragraph{Geometric deep learning}

Upon the success of Convolutional Neural Networks (CNNs) on Euclidean domains with grid-based structured data, the generalization of CNNs on non-Euclidean domains, \emph{e.g.}, graphs or meshes, has drawn significant attention~\citep{Bron2017}.
Therefore, a set of methods emerged, which aim to extend the convolution operator in deep neural networks to handle irregular structured input data by preserving its main properties: local connectivity, weight sharing and shift invariance~\citep{Lecun1998}.

\citet{Mas2015} introduced the first spatial mesh filtering approach by locally parameterizing the surface around each point using geodesic polar coordinates and defining convolutions on the resulting angular bins.
\citet{Bos2016} improved this approach by introducing a patch rotation method to align extracted patches based on the local principal curvatures of the input manifold.
\citet{Monti2017} proposed a method for continuous convolution kernels in the spatial domain by encoding spatial relations in edge attributes and using Gaussian kernels with trainable mean vector and covariance matrix as weight functions.
\citet{Fey2017} introduced SplineCNN with continuous kernel functions based on B-splines, which make it possible to directly learn from the geometric structure of meshes.

\subsection{Our contribution}

We present two graph based methods for the classification of cuneiform signs
represented by an original graph model. The first one compares two signs by the graph edit
distance with a tailored cost function in order to implement a nearest neighbor classifier.
The approach is able to take the relative relation between wedges into account and
thereby overcomes limitations of approaches based on optimal assignments, where
the costs are determined by the Euclidean distance in a global coordinate system.
We present efficient problem-specific heuristics for finding suboptimal solutions,
for which we show experimentally that they are sufficiently accurate for our application.

The second method employs the graph convolutional network SplineCNN.
We show that augmentation on the training set, that is applying random affine
transformation on vertex positions, yields significant improvements in accuracy.
SplineCNN provides fast inference algorithms, so that graph neural networks offer a real alternative.
We evaluate and compare both methods on a newly created, self-annotated graph based cuneiform dataset, which has been made publicly
available\footnote{\url{http://graphkernels.cs.tu-dortmund.de}} to encourage further research.

\section{Preliminaries}

We consider directed graphs $G=(V,E)$, where $V$ are the vertices and
$E \subseteq V \times V$ the edges. We allow vertices and edges to be
annotated by arbitrary attributes or features and denote the
$d$-dimensional position of the vertex $v_i$ by $\vec{p}_i \in \mathbb{R}^d$.
For a vertex $v_i \in V$ its neighborhood set is denoted by $\set{N}(v_i)$.
We briefly recall the fundamentals of the two graph based techniques used for
cuneiform recognition in the following.

\subsection{Graph edit distance}
The graph edit distance relies on the following elementary operations to edit a
graph:
vertex substitution, edge substitution, vertex deletion, edge deletion, vertex
insertion and edge insertion.
Each operation $o$ is assigned a cost $c(o)$, which may depend on the attributes
associated with the affected vertices and edges. A sequence of edit operations
$(o_1,\dots,o_k)$ that transforms a graph $G$ into another graph $H$ is called
an \emph{edit path} from $G$ to $H$. We denote the set of all possible edit
paths from $G$ to $H$ by $\Upsilon(G,H)$.
Let $G$ and $H$ be attributed graphs. The \emph{graph edit distance} from
$G$ to $H$ is defined by
\begin{equation}
d(G,H) = \min_{(o_1,\dots,o_k) \in \Upsilon(G,H)} \sum_{i=1}^k c(o_i).
\end{equation}
In order to obtain a meaningful measure of dissimilarity for graphs, a cost
function must be tailored to the particular attributes that are present in the
considered graphs.

\subsection{Graph convolutional network}

For the graph convolutional network, we use SplineCNNs with their continuous spline-based convolution operator by~\citet{Fey2017}.
SplineCNNs are a generalization of traditional CNNs for irregular structured data.
For a graph $G$ with $n$ vertices we store the $m$ vertex attributes (without the positions) in a feature matrix $\vec{F} \in \mathbb{R}^{n \times m}$.
Convolution over neighboring features for a vertex $v_i \in V$ is then defined by
\begin{equation}
  {(\vec{F} \star \vec{g})}_{i} = \frac{1}{|\set{N}(v_i)|} \sum_{l=1}^{m} \sum_{v_j \in \set{N}(v_i)} F_{j,l} \cdot g_l(\vec{p}_i - \vec{p}_j),
\end{equation}
where $\vec{g} =(g_1, \ldots, g_m)$ defines $m$ continuous kernel functions, which weight the components of $\vec{F}_{j}$ based on the spatial relation $\vec{p}_i - \vec{p}_j$ between vertex $v_i$ and $v_j$.
A kernel function $g_l$ is parameterized by a fixed number of trainable parameters.
For computing $g_l(\vec{p}_i - \vec{p}_j)$, the kernel function relies on the product of B-spline basis functions of a user-defined degree and can thus be evaluated very efficiently due to the local support property of B-splines.
While training, the parameters of the convolution operator are optimized using gradient descent.

\section{Representing cuneiform signs by graphs}%
\label{sec:dataset}

The dataset used in this work has been generated using the approach of~\citet{Fisseler2013} to automatically extract a tetrahedron based wedge model for individual wedges.
The selected cuneiform tablets have been chosen to guarantee the inclusion of a sufficient number of individual cuneiform signs and a sufficient number of representatives for each class.
As these requirements are difficult to meet on a small selection of ancient cuneiform fragments, we used a selection of nine cuneiform tablets written by scholars of Hittitology in the course of a study about individualistic characteristics of cuneiform hand writing.
The tablets were provided by the Hethitologie-Portal Mainz~\citep{Mueller2000}.
Each tablet contains up to 30 sylabic Hittite cuneiform signs, \emph{cf.}~\figureref{fig:tablet}, with each scribe contributing at least two tablets.
This results in a total of nine samples and four different scribes for each sign.
The signs were chosen to include either visually very distinctive wedge constellations and visually extremely similar signs.
Therefore, the dataset includes both challenges to test the generalization and the discriminization capabilities of the methods.
As the used segmentation approach extracts only individual wedge models, manual post-processing was applied to specify the affiliation of the wedges to the cuneiform signs.

\subsection{Graph model}

\begin{figure}[tb]
\floatconts%
  {fig:tablet_graph}
  {\caption{Cuneiform tablet and the graph representation of sign 2.}}
  {%
    \null\hfill
    \subfigure[Cuneiform tablet]{\label{fig:tablet}%
       \includegraphics[height=4cm]{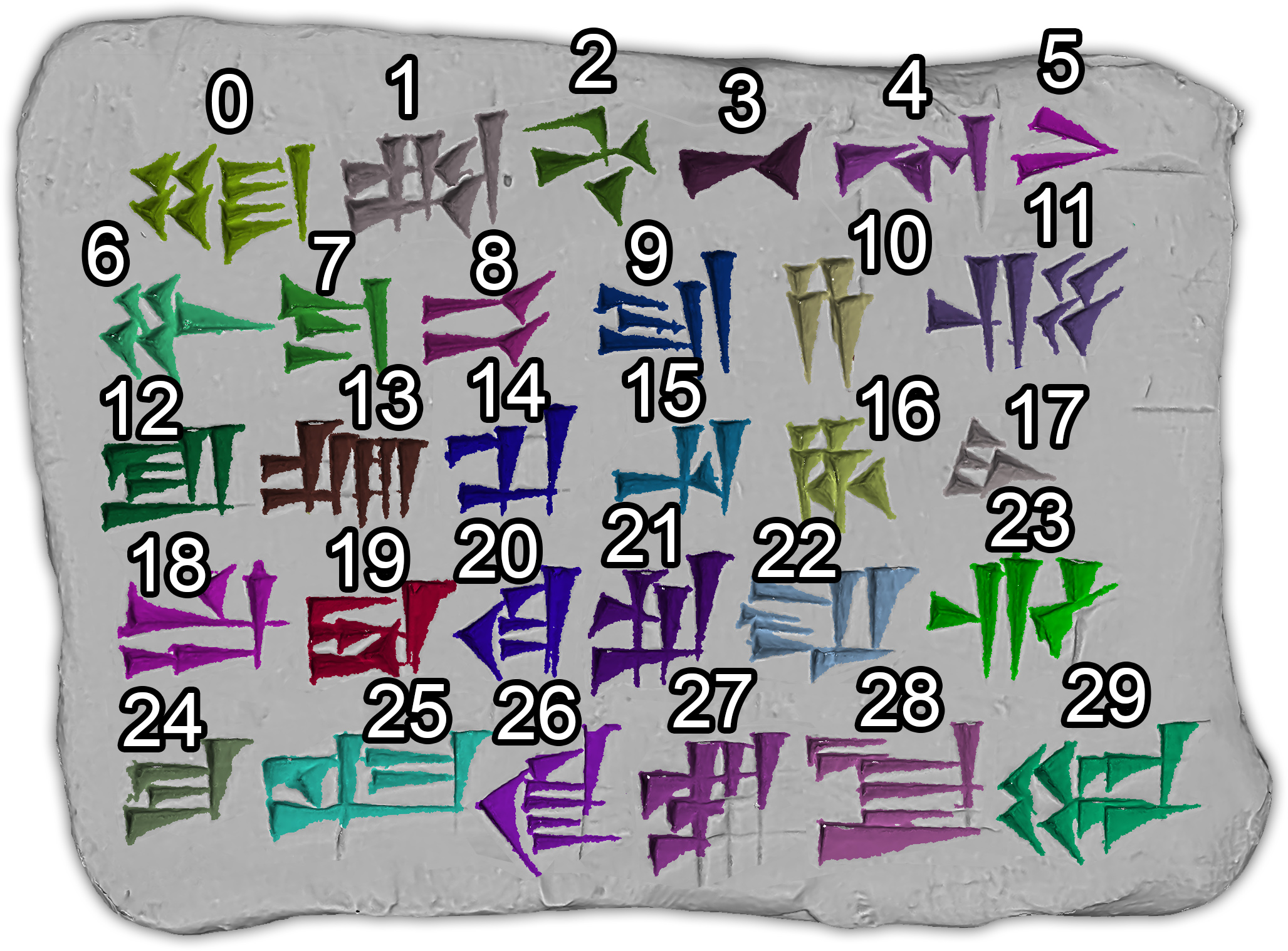}%
    }%
    \hfill
    \subfigure[Graph representation]{\label{fig:graph}%
      \includegraphics[height=4cm]{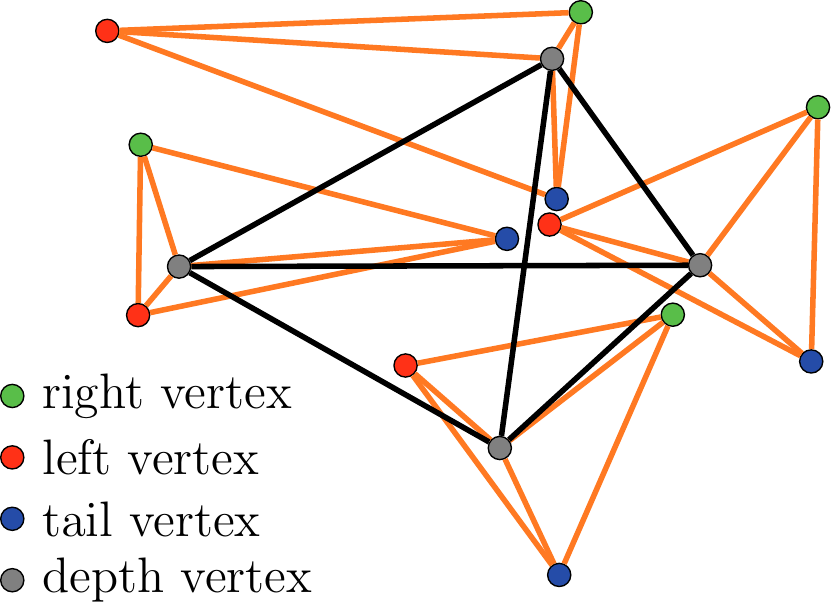}%
    }%
    \hfill\null%
  }
\end{figure}

We represent each wedge by four vertices, which are labeled by their point (depth, tail, right, left) respectively glyph types (vertical, horizontal, `Winkelhaken') and their spatial coordinates in $\mathbb{R}^2$.
The vertices of the same wedge are pairwise adjacent and form a clique, as shown in \figureref{fig:graph}.
The relative relation between individual wedges is not adequately modeled by their
location in the global coordinate system. In \figureref{fig:tablet} the signs
`tu' (number 0) and `li' (number 29) differ only by the arrangement of the four
horizontal wedges, which are either ordered from top to bottom or arranged in a
square. While this difference is easily recognized visually, the one sign can
be turned into the other without shifting wedges far away.
In order to obtain connected graphs we introduce additional edges between all 
pairs of depth vertices.
These edges are also used to compare the relative positions of wedges and
we refer to them as \emph{arrangement edges}.
In the following section we describe how arrangement edges are matched using the
graph edit distance.
Our experimental evaluation in \sectionref{sec:exp} will show that these edges
are crucial to obtain a meaningful distance in classification tasks for some
cuneiform signs.

\section{Comparing cuneiform signs using the graph edit distance}\label{sec:ged}
In order to obtain a meaningful distance measure for cuneiform graphs we employ
the following cost function.
The vertex substitution costs are $\infty$ for vertices with different
labels (vertex type or glyph type) and the squared Euclidean distance between
the associated coordinates otherwise.
The costs for substituting arrangement edges are chosen to reflect the consistency of 
the relative relations between the associated wedges. Therefore, we annotate 
the arrangement edge $(v_i, v_j)$ by the Euclidean vector $\vec{p}_j-\vec{p}_i$.
The cost for substituting an arrangement edge annotated by $\vec{x}$ for
an arrangement edge annotated by $\vec{y}$ are then determined by the Cosine
distance as \mbox{$1 - \vec{x}^\top\vec{y}/(\norm{\vec{x}}\norm{\vec{y}})$}
weighted by a parameter $\alpha$ in order to balance the influence of vertex
and edge substitutions.
The other edges are substituted at no cost; we forbid to substitute edges of
different type by setting the cost function to $\infty$ in this case.
In order to support error-tolerant graph matching, we allow vertex and
edge deletion/insertion at a (typically high) cost of $D$.
We exploit the characteristics of this cost function in order to find efficient
heuristic solutions in the following.

\subsection{Heuristics for computing the cuneiform graph edit distance}\label{sec:ged:apx}
We present efficient heuristics for computing the edit distance between two cuneiform
graphs. \citet{Riesen2009a} proposed to derive a suboptimal edit path between
graphs from an optimal assignment of their vertices, where the assignment costs
encode the local edge structure.

Considering our graph model and cost function, we observe that the four vertices
representing the same wedge in one graph must be mapped to the corresponding
vertices of one wedge in the other graph to avoid high costs for deleting edges.
This motivates to compute an optimal assignment between wedges to obtain an edit
path. For computing the assignment we initially ignore the costs caused by
substituting arrangement edges, but take them into account when computing the
cost of the edit path derived from the assignment.
In more detail, we proceed as follows: For two cuneiform graphs $G$ and $H$ with
wedges $\set{A} = \{a_1,\dots,a_n\}$ and $\set{B} = \{b_1,\dots,b_m\}$,
we create the assignment cost matrix $\vec{C}$ according to
\begin{equation}
\vec{C}=
 \begin{bmatrix}[cccc|cccc]
c_{1,1} & c_{1,2} & \cdots & c_{1,m} & c_{1,\epsilon} & \infty & \cdots  & \infty \\
c_{2,1} & c_{2,2} & \cdots & c_{2,m} & \infty & c_{2,\epsilon} & \ddots & \vdots \\
\vdots & \vdots & \ddots & \vdots & \vdots & \ddots & \ddots &\infty \\
c_{n,1} & c_{n,2} & \cdots & c_{n,m} & \infty & \cdots & \infty  & c_{n,\epsilon} \\

\addlinespace[-0.2\aboverulesep]\cmidrule[0.4pt](l{4pt}r{4pt}){1-8}\addlinespace[-1.0\belowrulesep]

c_{\epsilon,1} & \infty & \cdots & \infty & 0 & 0 & \cdots & 0 \\
\infty & c_{\epsilon,2} & \ddots & \vdots & 0 & 0 & \ddots & \vdots \\
\vdots & \ddots & \ddots & \infty & \vdots & \ddots & \ddots & 0 \\
\infty & \cdots & \infty & c_{\epsilon,m} & 0 & \cdots & 0  & 0 \\
\end{bmatrix},
\end{equation}
where $c_{i,j}$ denotes the cost of assigning the wedge $a_i$ to the wedge $b_j$
of the other sign, $c_{i,\epsilon}$ is the cost of deleting the wedge $a_i$ and
$c_{\epsilon,j}$ the cost of inserting the wedge $b_j$.
The cost for assigning two wedges is the sum of squared Euclidean distances
between their vertices of the same type, if the wedges are of the same glyph
type, and $\infty$ otherwise.
Since the deletion (insertion) of a wedge involves at least the deletion (insertion)
of four vertices and twelve directed edges, we set
$c_{i,\epsilon}=c_{\epsilon,j}=16D$ for $i \in \{1,\dots,n\}$ and
$j \in \{1,\dots,m\}$.
It should be noted that the costs defined above do not take the deletion, insertion and substitution costs of arrangement edges into account.

We find the optimal assignment for the problem defined above in cubic time using
the Hungarian method. From this approach we derive two approximation algorithms:
First, we directly use the optimal assignment cost as an approximation of
the graph edit distance (APX1). This approach ignores all costs caused
by arrangement edges.
The second approach (APX2) determines the mapping of vertices of $G$ and $H$ from the
assignment of wedges and determines the associated edit path, which includes
the substitution of arrangement edges. With this approach the exact cost of
this (not necessarily optimal) edit path is used as an approximation of the
graph edit distance. Comparing APX1 to APX2 allows us to evaluate the importance
of the arrangement edges experimentally.

\section{Classifying cuneiform graphs using graph convolutional networks}

The used graph convolutional neural network takes 8-dimensional input features for each vertex and outputs a confidence distribution over the different cuneiform signs.
Input features for a vertex are given by its point type (depth, tail, right and left) and glyph type (vertical, horizontal, `Winkelhaken') as either $-1$ or $1$ as well as an additional $1$ for each vertex to be able to learn from the general geometry of cuneiform signs itself.

\figureref{fig:network} gives a comprehensive view of the used network architecture.
The network consists of three convolutional layers and therefore aggregates local $3$-neighborhoods.
We use Exponential Linear Units (ELUs) as non-linearities after each layer.
In each layer, we extend the number of feature maps from 8 to 32 to finally 64 maps.
For the spline-based convolution operator by~\citet{Fey2017}, we use a B-spline with degree of 1, which is parameterized by 25 trainable parameters.
After that, we average features in the vertex dimension, which enables the use of a fully-connected layer from 64 to 30 neurons.
Applying softmax, we obtain a confidence distribution over 30 different cuneiform signs.

\begin{figure}[tb]
\floatconts%
  {fig:network}
  {\caption{Graph convolutional network architecture for classifying 30 cuneiform signs.}}
  {%
    \includegraphics[width=\linewidth]{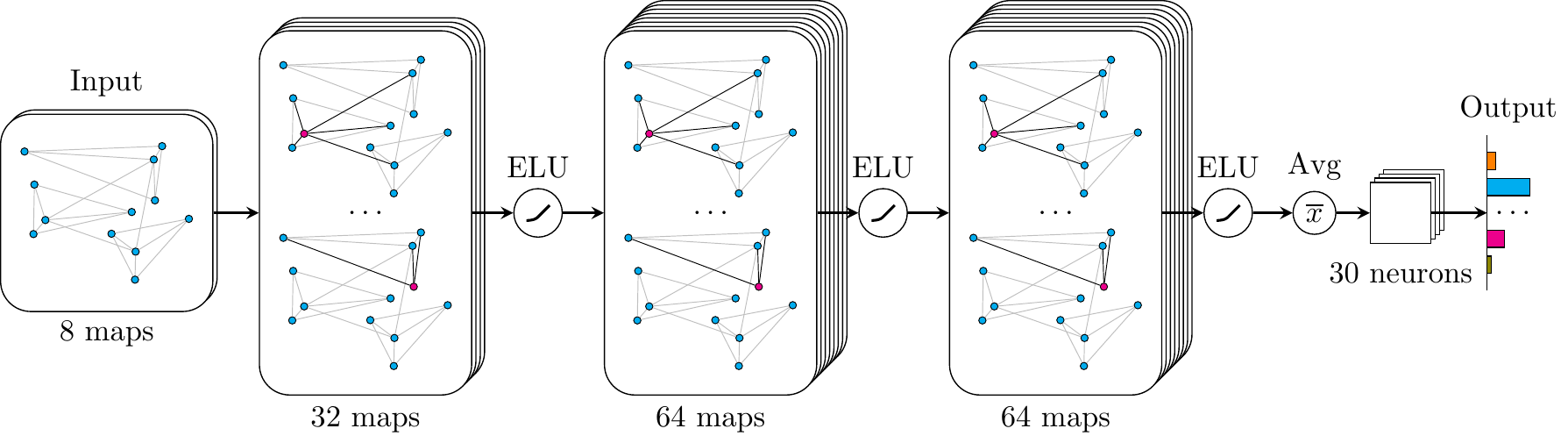}
  }
\end{figure}

\subsection{Augmentation}

We make heavy use of augmentation on the training set in order to reduce overfitting by applying random affine transformations on the graph vertices respectively graphs.
Translation, scaling and rotation on a graph vertex with position $\vec{p} \in \mathbb{R}^2$ are therefore given by
\begin{equation}
\overline{\vec{p}} = \vec{p} + \begin{pmatrix}t_1\\t_2\end{pmatrix},
\quad
\overline{\vec{p}} = \begin{pmatrix}s_1&0\\0&s_2\end{pmatrix} \cdot \vec{p},
\quad
\overline{\vec{p}} = \begin{pmatrix}\cos(\theta)&-\sin(\theta)\\\sin(\theta)&\phantom{-}\cos(\theta)\end{pmatrix} \cdot \vec{p},
\end{equation}
where $t_1, t_2 \in [-T, T]$, $s_1, s_2 \in [\frac{1}{S}, S]$ and $\theta \in [-\Theta, \Theta]$ are randomly sampled within a given boundary.
It should be noted that affine transformations are not commutative and one might also consider randomizing their order of application.
Here, we use a fixed order by first rotating and scaling the whole graph and finally translating individual graph vertices.

\section{Experimental evaluation}\label{sec:exp}

We are interested in answering the following questions:
\begin{description}[align=right,labelindent=!,topsep=4pt,itemsep=1ex,parsep=0ex]
\item[Q1] Is the cuneiform graph edit distance with the suggested cost function
          a meaningful measure to compare cuneiform signs?
          How do the heuristics for computing the graph edit distance compare
          to exact methods in terms of runtime and solution quality?
\item[Q2] How do CNNs perform in comparison to a nearest neighbor classifier
          using the cuneiform graph edit distance?
          How sensitive is the SplineCNN to the training set size?
\item[Q3] How do both methods compare in terms of runtime?
          What is the impact of the training set size in practice?
\end{description}

\subsection{Experimental setup}

For all experiments, we used the dataset described in \sectionref{sec:dataset}, which contains $267$ cuneiform signs with 5,680 vertices and 23,922 edges in total.
A comprehensive overview of the dataset can be found in \tableref{tab:dataset}.

\begin{table}[htbp]
\floatconts%
{tab:dataset}
{\caption{%
  Dataset characteristics for all 30 classes: Number of graphs $\#G$, vertices $|V|$, edges $|E|$ and the maximum height $h_{\max}$ and width $w_{\max}$ of the axis-aligned minimum bounding boxes.}}
{\resizebox{\textwidth}{!}{%
\begin{tabular}{lrrrrrrrrrrrrrrr}
\toprule
Class & `ba' & `bi' & `bu' & `da' & `di' & `du' & `ha' & `hi' & `hu' & `ka' & `ki' & `ku' & `la' & `li' & `lu' \\
\midrule
$\#G$ & 9 & 9 & 9 & 9 & 9 & 9 & 9 & 9 & 9 & 9 & 9 & 9 & 9 & 9 & 9 \\
$|V|$ & 36 & 24 & 16 &  8 & 16 &  8 & 20 & 16 & 16 & 20 & 16 & 28 & 24 & 28 & 16 \\
$|E|$ & 180 & 102 & 60 & 26 & 60 & 26 & 80 &  60 & 60 & 80 &  60 & 126 & 102 & 126 &  60 \\
$h_{\max}$ & 7.5 & 7.4 & 10.0 & 5.3 & 7.5 & 7.7 &  7.3 & 7.9 & 7.5 & 9.1 &  9.1 &  7.4 & 7.7 & 6.7 &  7.5 \\
$w_{\max}$ & 12.7 & 11.7 & 13.4 & 7.3 & 14.7 &  8.7 & 13.9 & 10.0 &  9.4 & 11.1 &  5.8 & 13.7 & 11.4 & 10.6 & 15.0 \\
\midrule
Class & `na' & `ni' & `nu' & `ra' & `ri' & `ru' & `sa' & `si' & `su' & `ta' & `ti' & `tu' & `za' & `zi' & `zu' \\
\midrule
$\#G$ & 9 & 9 & 9 & 9 & 9 & 9 & 9 & 9 & 9 & 9 & 9 & 9 & 8 & 8 & 8 \\
$|V|$ & 16 & 24 & 16 & 20 & 20 & 20 & 20 & 24 & 20 & 16 & 28 & 28 & 24 & 28 & 36 \\
$|E|$ &  60 & 102 & 60 & 80 & 80 & 80 & 80 & 102 & 80 & 60 & 126 & 126 & 102 & 126 & 180 \\
$h_{\max}$ & 8.5 & 9.9 & 10.2 & 7.7 & 8.9 & 8.2 & 10.0 & 9.2 & 8.2 & 8.3 & 12.6 & 10.5 & 9.1 & 8.9 & 10.3 \\
$w_{\max}$ & 14.4 &  7.6 &  8.6 & 9.6 & 11.5 & 10.2 & 10.4 & 11.2 & 11.7 &  8.4 & 15.2 & 12.4 &  9.2 & 13.2 & 21.0 \\
\bottomrule
\end{tabular}}}
\end{table}

\paragraph{Edit distance}
We implemented an exact method for computing the graph edit distance based on
a recent binary linear programming formulation~\citep{Lerouge2017} and solved
all instances using Gurobi v7.5.2. The exact method as well as the two heuristics
were implemented in Java. The experiments were conducted using Java v1.8.0
on an Intel Core i7-3770 CPU at 3.4GHz with 16GB of RAM using a single processor
only.
We set the parameters $\alpha=1000$ and $D=1000$.

\paragraph{Graph convolutional network}

We trained the proposed SplineCNN architecture\footnote{\url{https://github.com/rusty1s/pytorch_geometric}} on a single NVIDIA GTX 1080 Ti with 11GB of memory using PyTorch.
Each experiment was trained for 300 epochs using Adam optimization~\citep{Kingma2015} and cross entropy loss, with a batch size of 32, dropout probability of 0.5, an initial learning rate of 0.01 and learning rate decay to 0.001 after 200 epochs.
Hyperparameters for augmentation were set to $T=0.1$, $S=1.4$ and $\Theta=0.6$.

\subsection{Evaluation of the cuneiform graph edit distance (Q1)}

We compared the exact method and the two heuristics APX1 and APX2 described in
\sectionref{sec:ged:apx}.
We used the signs of one tablet as a reference and, for each reference
sign, sorted all the elements of the dataset according to their distance to the
reference sign. We expect that the other instances of the same sign have a small
distance and populate the top ranks. We consider a classifier which assigns the
top $k$ elements in the ranking to the same class as the reference sign and
considered the ROC curves obtained for varying $k$.

\begin{figure}[tb]
\floatconts%
  {fig:roc_selection}
  {\caption{Selected ROC curves for the ranking experiments.}}
  {%
   \subfigure[Sign `ba']{\label{fig:roc_selection:ba}
      \includegraphics[scale=.8]{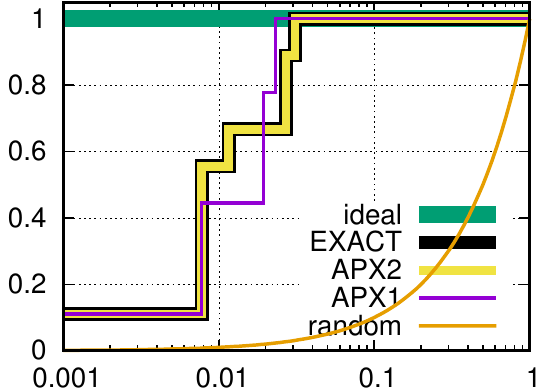}
   }\hfill
   \subfigure[Sign `ki']{\label{fig:roc_selection:ki}
      \includegraphics[scale=.8]{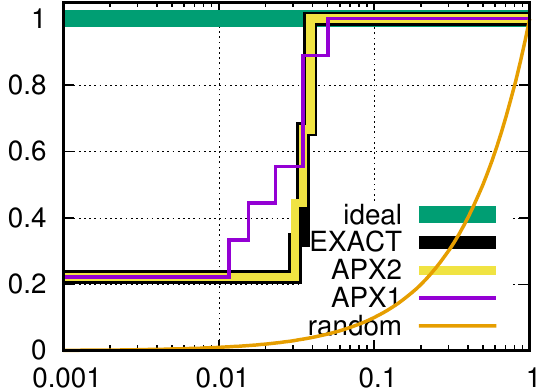}
   }\hfill
   \subfigure[Sign `ku']{\label{fig:roc_selection:ku}
      \includegraphics[scale=.8]{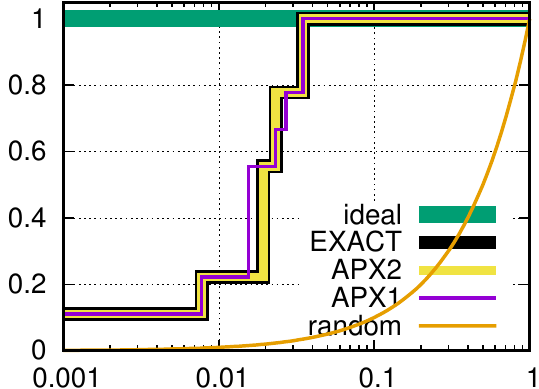}
   }\hfill
   \subfigure[Sign `li']{\label{fig:roc_selection:li}
      \includegraphics[scale=.8]{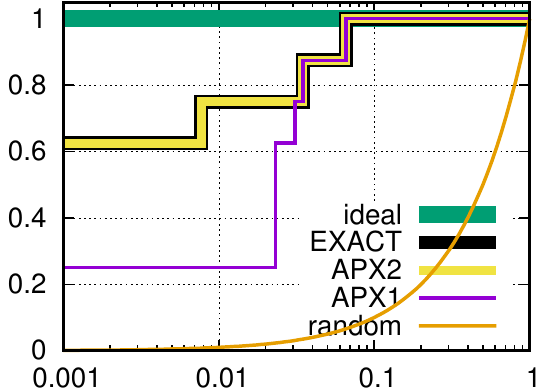}
   }\hfill
   \subfigure[Sign `ni']{\label{fig:roc_selection:ni}
      \includegraphics[scale=.8]{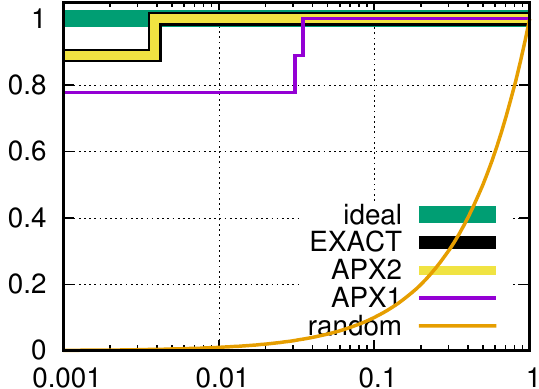}
   }\hfill
   \subfigure[Sign `tu']{\label{fig:roc_selection:tu}
      \includegraphics[scale=.8]{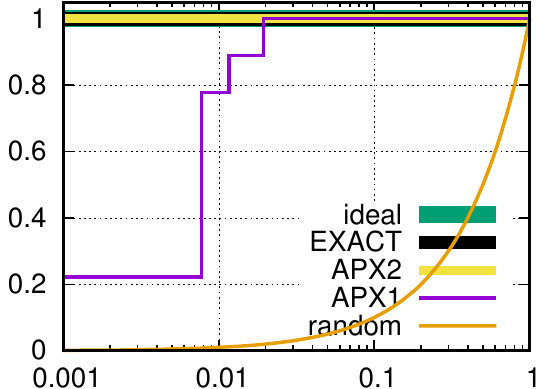}
   }}
\end{figure}

We obtained ideal results  (AUC 1) for 16 of 30 signs by all three methods, \emph{i.e.},
most signs are perfectly recognized by APX1, APX2 and the exact method.
\figureref{fig:roc_selection} shows a selection of the resulting ROC curves for
more challenging signs, see Appendix~\ref{apd:roc} for a complete overview.
Please note the logarithmic scale of the $x$-axis.
\figureref{fig:roc_selection:ba} and \figureref{fig:roc_selection:ku} show a
clear deviation from the ideal. The reason for this is that the signs `ba' and
`ku' are often not correctly distinguished and have a small distance.
On a purely geometric level without semantic context, these signs are difficult
to distinguish even for human experts, see signs number 7 and 24 in \figureref{fig:tablet}.
APX1 and APX2 both perform well in most cases. However, in order to distinguish
`li' and `tu' the relative relation between wedges has to be taken into account,
see signs number 0 and 29 in \figureref{fig:tablet}.
For these two signs we observe that APX2 indeed outperforms APX1, \emph{cf.}\@
\figureref{fig:roc_selection:li} and \figureref{fig:roc_selection:tu}.
The poor performance for the sign `ki', \emph{cf.}\@ \figureref{fig:roc_selection:ki}, can be
explained by the fact that the reference sign contains a scribal error on the
reference tablet. \figureref{fig:roc_selection:ni} is a typical example showing
a satisfying performance, where only very few instances of the reference signs
were ranked slightly below other signs.
In almost all experiments we observed that APX2 leads to the same ROC curve as
the exact method. Moreover, we verified that both compute the same distance with
very few exceptions primarily for examples widely separated.

The average runtime for computing the rankings for all 30 signs, which involves
more than 80,000 distance computations, was 27.27 minutes for the exact method,
48.7 milliseconds for APX1 and 53.7 milliseconds for APX2.
In summary, we have shown that the cuneiform graph edit distance provides an adequate
measure for the comparison and recognition of cuneiform signs and can be
efficiently approximated by APX2 without sacrificing accuracy.

\subsection{Classification via edit distance and SplineCNN (Q2)}

We performed $10$-fold cross validation on random splits of the dataset in order to provide a fair comparison.
For SplineCNN, the accuracy for each split is given by the average over 10 experiments due to random uniform weight initialization.
For the cuneiform graph edit distance we used a $3$-nearest neighbors classifier on the same splits.
The accuracy results are summarized in Table~\ref{tab:results}.

\begin{table}[tb]
\floatconts%
  {tab:results}
  {\caption{Cuneiform sign classification results of all proposed methods for each test split, overall mean accuracy and standard deviation.}}
  {\resizebox{\textwidth}{!}{%
  \begin{tabular}{lrrrrrrrrrrrr}
  \toprule
  & & & \multicolumn{9}{c}{\# Split} \\
  \cmidrule{3-12}
  Method & Mean & 1 & 2 & 3 & 4 & 5 & 6 & 7 & 8 & 9 & 10 \\
  \midrule
  GED EXACT & 93.24 & \textbf{96.30} & \textbf{100.0} & 96.30 & \textbf{100.0} & 92.59 & 85.19 & 85.19 & 85.19 & \textbf{100.0} & 91.67 \\
            & $\pm$ 5.94&   &       &       &       &       &       &       &       &       &       \\
  GED APX1  & 89.17 & 92.59 & 96.30 & 96.30 & 88.89 & 85.19 & 81.48 & 81.48 & 81.48 & 96.30 & 91.67 \\
            & $\pm$ 6.03&   &       &       &       &       &       &       &       &       &       \\
  GED APX2  & 92.87 & 92.59 & \textbf{100.0} & 96.30 & \textbf{100.0} & 92.59 & 85.19 & 85.19 & 85.19 & \textbf{100.0} & 91.67 \\
            & $\pm$ 5.86&   &       &       &       &       &       &       &       &       &       \\
  \midrule
    CNN without & 87.37 & 86.56 & 86.30 & 87.41 & 90.74 & 92.60 & \textbf{93.33} & 87.04 & 79.26 & 81.48 & 90.00 \\
  augmentation & $\pm$ 6.16 & $\pm$ 4.43 & $\pm$ 3.92 & $\pm$ 5.84 & $\pm$ 6.11 & $\pm$ 4.62 & $\pm$ 2.92 & $\pm$ 3.15 & $\pm$ 5.00 & $\pm$ 3.90 & $\pm$ 4.89 \\
    CNN with & \textbf{93.54} & 95.92 & 94.81 & \textbf{100.0} & 94.44 & \textbf{95.19} & 89.63 & \textbf{90.74} & \textbf{88.89} & 99.37 & \textbf{95.42} \\
  augmentation & $\pm$ 4.40 & $\pm$ 2.73 & $\pm$ 2.59 & $\pm$ 0.00 & $\pm$ 3.15 & $\pm$ 1.79 & $\pm$ 2.92 & $\pm$ 4.00 & $\pm$ 4.62 & $\pm$ 1.91 & $\pm$ 3.65 \\
  \bottomrule
  \end{tabular}}}
\end{table}

The graph edit distance based classifiers provide a high classification accuracy.
The exact method and APX2 both outperform APX1, which can be explained by the impact of arrangement edges.
For SplineCNN, we obtain high accuracies, although deep learning is well known for demanding a huge amount of training data.
We can even beat the proposed approaches by a small margin with the use of augmentation, that is generating augmented examples in each epoch, which differ from the training data by a random affine transformation.
However, choosing a smaller amount of training data yields lower results due to an decrease in training and an increase in test data size.
For example, on a pre-chosen 50/50 split, we obtain an accuracy of 90.23\%, resulting in a quite sensitive approach to the training set size.
Moreover, it is worth pointing out that the output accuracies of different test splits can yield quite varying results.
This can be traced back to the small number of test graphs, approximately 27, where a single false classification leads to jumps in accuracy of approximately 3.70 percentage points.

It can be further observed that there exist a few test splits, where the exact method and APX2 both outperform APX1 and SplineCNN, \emph{cf.} split 4.
This indicates that SplineCNN faces difficulties identifying characters where arrangement edges play a crucial role.
This behaviour \emph{might be} explained by the uniform knot vectors of the B-spline kernels, as the kernel functions lose granular control for neighborhoods with very long as well as very short edges.

In summary, all methods provide a high classification accuracy with the SplineCNN performing best when the training set is augmented.

\subsection{Evaluation of runtimes (Q3)}

We experimentally evaluated the runtime of both methods for training and
testing with varying dataset sizes. We report the runtimes averaged over
10 runs for SplineCNN, APX1 and APX2. For the nearest neighbor classifier no training is required and we
report the prediction time for a fixed test set using training sets of varying size.
\tableref{tab:runtime} summarizes the results.
The runtimes generally grow linearly with the number of graphs in the training and
test set.
For SplineCNN, the runtime to train all 267 cuneiform signs is 6.7
seconds, whereas the runtime for classifying all signs is very fast with approximately
14 milliseconds. Classification using the exact method is very slow in practice,
while APX1 and APX2 both are fast for our dataset. However, since their
prediction time increases with the training set size these approaches
can be expected to become less feasible when the available training data grows.
\begin{table}[tb]
\floatconts%
 {tab:runtime}%
 {\caption{%
   Runtime analysis of the used SplineCNN architecture for varying training and test sizes ($a$)
   and for classifying 50\% of the signs by the graph distance based nearest neighbor classifier
   for varying training set sizes ($b$).}}%
 {%
   \subtable[SplineCNN]{%
     \label{tab:runtime_cnn}%
     \resizebox{0.385\textwidth}{!}{%
     \begin{tabular}{lrr}
     \toprule
       Size [\%] & Training [s] & Test [ms] \\
      \midrule
      25 & $1.95 \pm 0.02$ & $3.56 \pm 0.18$   \\
      50 & $3.51 \pm 0.03$ & $7.31 \pm 0.49$   \\
      75 & $5.08 \pm 0.01$ & $11.46 \pm 0.89$  \\
      100 & $6.66 \pm 0.03$ & $13.99 \pm 0.39$ \\
     \bottomrule
     \end{tabular}}
   }\qquad
   \subtable[GED]{%
     \label{tab:runtime_ged}%
     \resizebox{0.535\textwidth}{!}{%
     \begin{tabular}{lrrr}
      \toprule
       Size [\%] & EXACT [s] & APX1 [ms] & APX2 [ms] \\
     \midrule
     25  & 1301.1 &  $52.8 \pm 1.17$ &  $56.5 \pm 2.94$ \\
     50  & 2617.3 & $108.2 \pm 11.4$ & $110.6 \pm 1.74$ \\
     75  & 4008.3 & $153.1 \pm 2.43$ & $165.4 \pm 5.04$ \\
     100 & 5407.8 & $204.5 \pm 2.42$ & $222.2 \pm 8.36$ \\
      \bottomrule
     \end{tabular}}
   }
 }
\end{table}

\section{Conclusion and future work}
We have proposed two complementary graph based methods for the recognition of 
cuneiform signs, which yield a first step towards supporting philologists in 
their analysis.
We have evaluated our methods on a newly created real-world dataset
w.r.t.\@ runtime and classification accuracy.
While both methods obtain high accuracy and efficiency, CNN based approaches 
might be advantageous when the available training data grows.

For future work, we would like to investigate our methods in the presence of 
scanning errors and extend our dataset by cuneiform signs that can 
only be distinguished in context.
For neural networks, a traditional approach for recognizing those would be the 
addition of recurrent modules.

\acks{This research was supported by the German Science Foundation (DFG) within the
Collaborative Research Center SFB 876 `Providing Information by
Resource-Constrained Data Analysis', project A6.
We thank our colleges from the Hethitologie-Portal Mainz~\citep{Mueller2000} for providing the scanned cuneiform tablets and for supporting us with their philological expertise.
We are also thankful to Jan Eric Lenssen and Pascal Libuschewski for proofreading and helpful advice.}

\bibliography{cuneiformClassification}

\newpage

\appendix

\section{ROC curves for rankings based on the graph edit distance}\label{apd:roc}
\begin{figure}[htbp]
\floatconts%
{fig:roc1}
{\caption{ROC curves for the ranking experiments (1/2).}}{%
\subfigure[Sign `ba']{%
   \includegraphics[scale=.77]{roc_ba}
}\hfill
\subfigure[Sign `bi']{%
   \includegraphics[scale=.77]{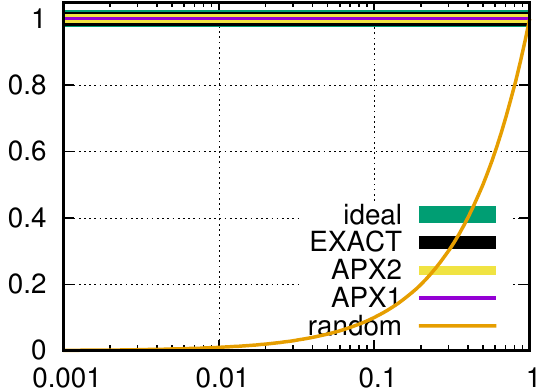}
}\hfill
\subfigure[Sign `bu']{%
   \includegraphics[scale=.77]{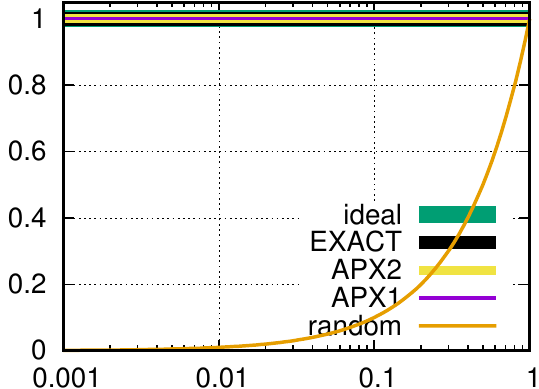}
}\hfill
\subfigure[Sign `da']{%
   \includegraphics[scale=.77]{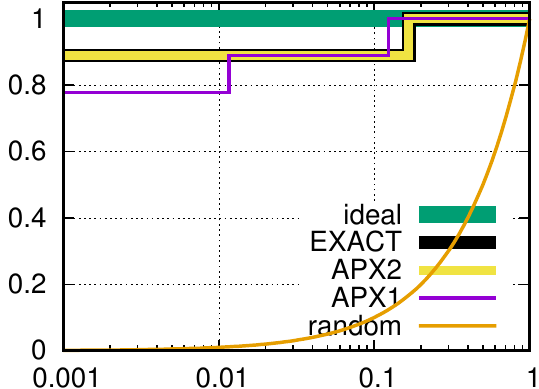}
}\hfill
\subfigure[Sign `di']{%
   \includegraphics[scale=.77]{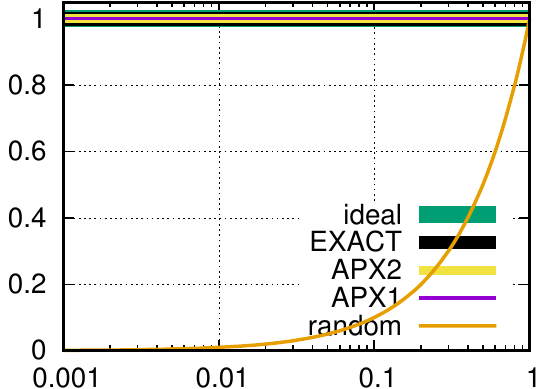}
}\hfill
\subfigure[Sign `du']{%
   \includegraphics[scale=.77]{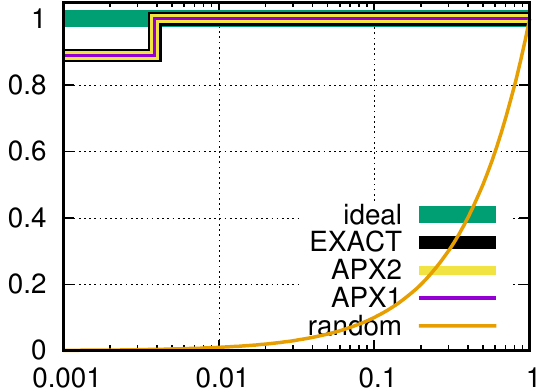}
}\hfill
\subfigure[Sign `ha']{%
   \includegraphics[scale=.77]{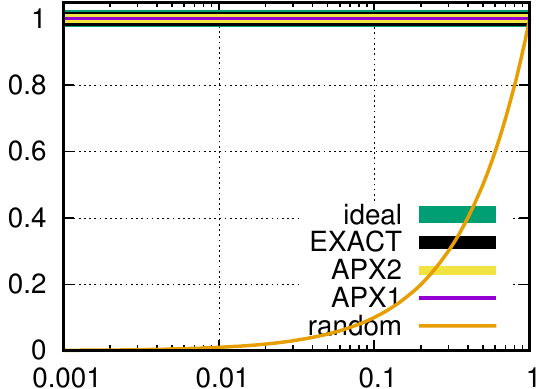}
}\hfill
\subfigure[Sign `hi']{%
   \includegraphics[scale=.77]{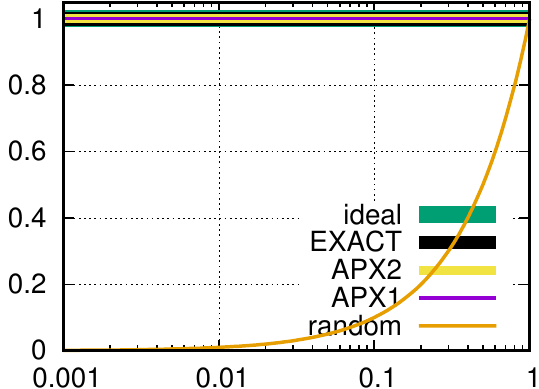}
}\hfill
\subfigure[Sign `hu']{%
   \includegraphics[scale=.77]{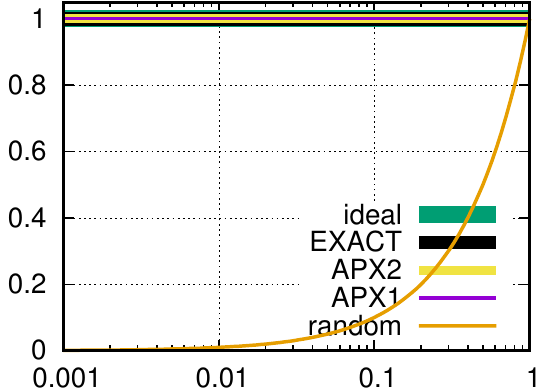}
}\hfill
\subfigure[Sign `ka']{%
   \includegraphics[scale=.77]{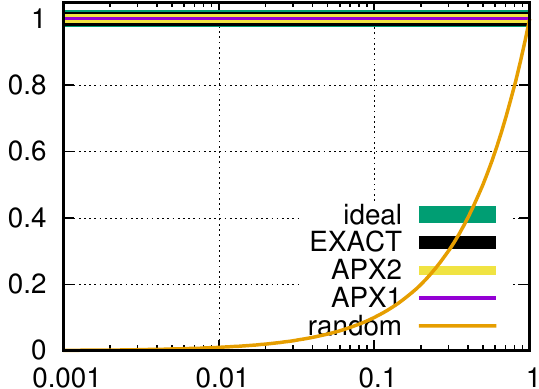}
}\hfill
\subfigure[Sign `ki']{%
   \includegraphics[scale=.77]{roc_ki}
}\hfill
\subfigure[Sign `ku']{%
   \includegraphics[scale=.77]{roc_ku}
}\hfill
\subfigure[Sign `la']{%
   \includegraphics[scale=.77]{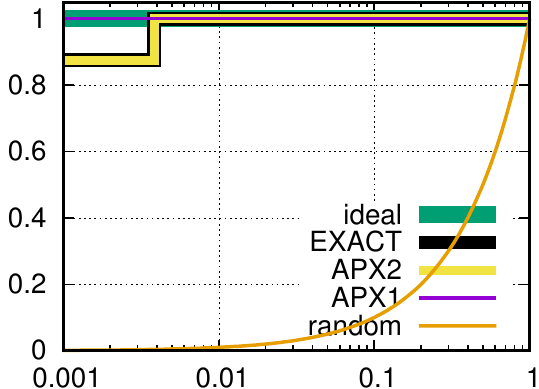}
}\hfill
\subfigure[Sign `li']{%
   \includegraphics[scale=.77]{roc_li}
}\hfill
\subfigure[Sign `lu']{%
   \includegraphics[scale=.77]{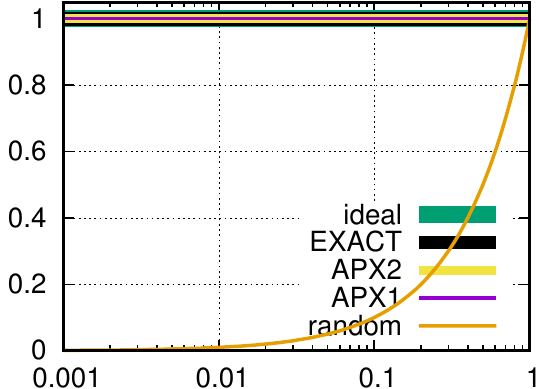}
}
}
\end{figure}
\begin{figure}[htbp]
\floatconts%
{fig:roc2}
{\caption{ROC curves for the ranking experiments (2/2).}}{%
   \subfigure[Sign `na']{%
      \includegraphics[scale=.77]{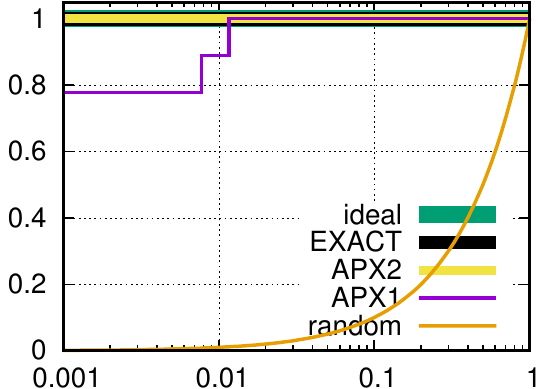}
   }\hfill
   \subfigure[Sign `ni']{%
      \includegraphics[scale=.77]{roc_ni}
   }\hfill
   \subfigure[Sign `nu']{%
      \includegraphics[scale=.77]{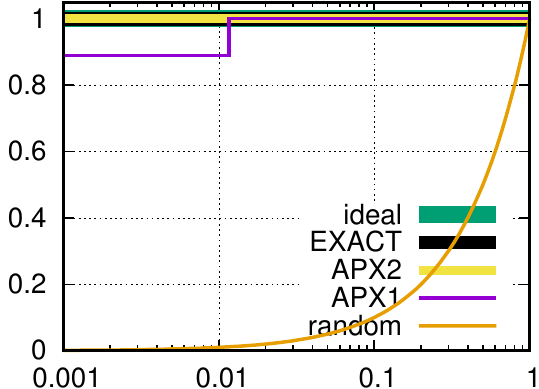}
   }\hfill
   \subfigure[Sign `ra']{%
      \includegraphics[scale=.77]{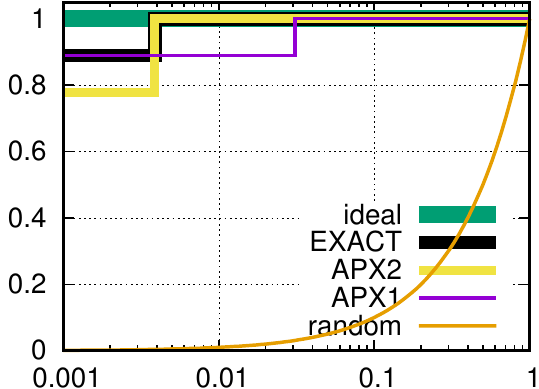}
   }\hfill
   \subfigure[Sign `ri']{%
      \includegraphics[scale=.77]{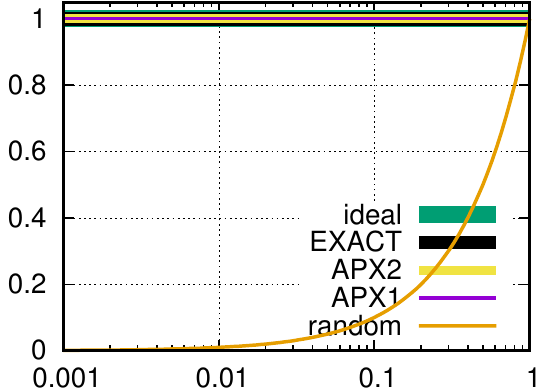}
   }\hfill
   \subfigure[Sign `ru']{%
      \includegraphics[scale=.77]{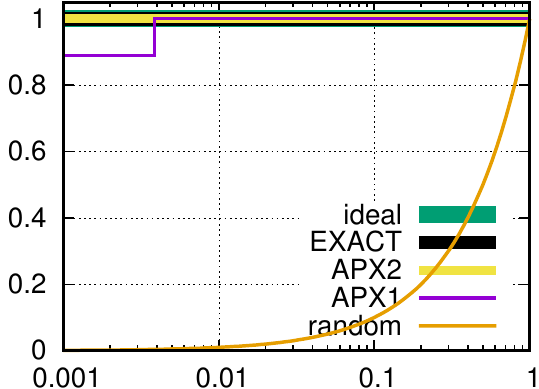}
   }\hfill
   \subfigure[Sign `sa']{%
      \includegraphics[scale=.77]{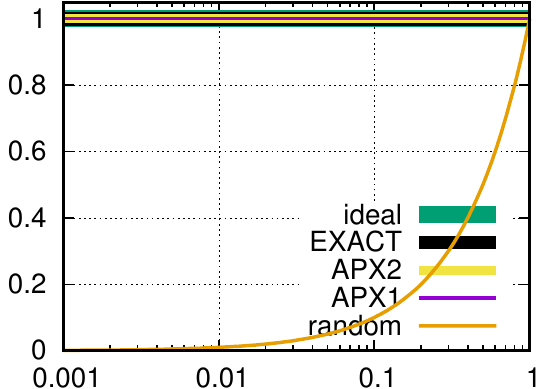}
   }\hfill
   \subfigure[Sign `si']{%
      \includegraphics[scale=.77]{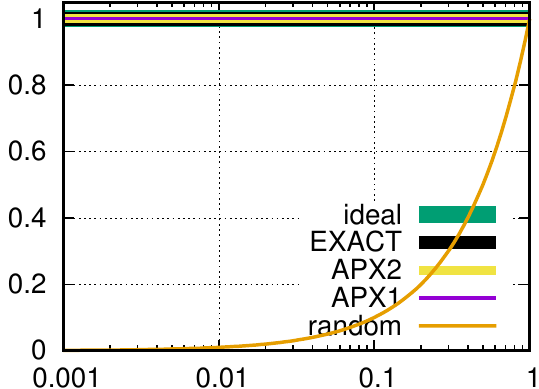}
   }\hfill
   \subfigure[Sign `su']{%
      \includegraphics[scale=.77]{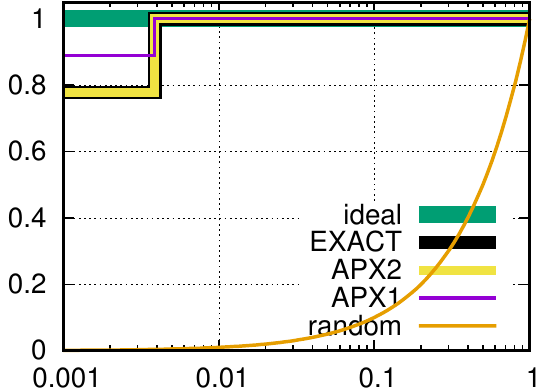}
   }\hfill
   \subfigure[Sign `ta']{%
      \includegraphics[scale=.77]{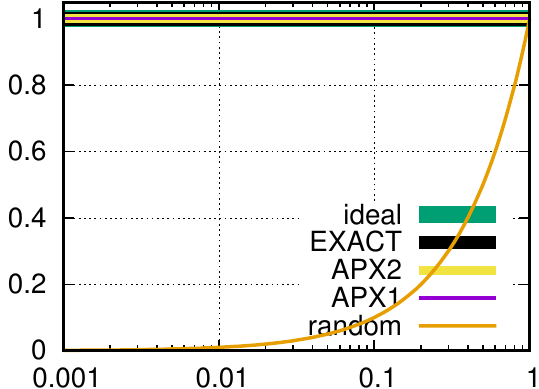}
   }\hfill
   \subfigure[Sign `ti']{%
      \includegraphics[scale=.77]{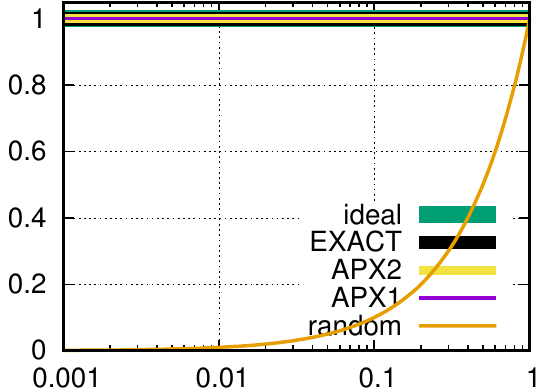}
   }\hfill
   \subfigure[Sign `tu']{%
      \includegraphics[scale=.77]{roc_tu}
   }\hfill
   \subfigure[Sign `za']{%
      \includegraphics[scale=.77]{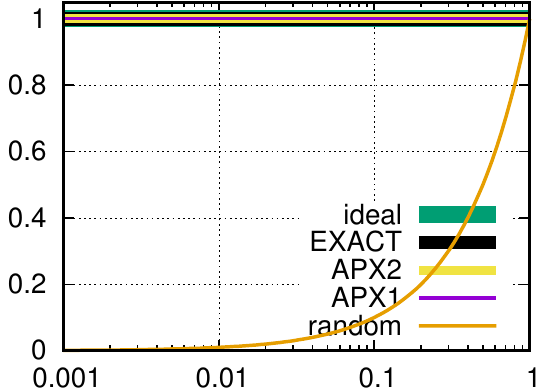}
   }\hfill
   \subfigure[Sign `zi']{%
      \includegraphics[scale=.77]{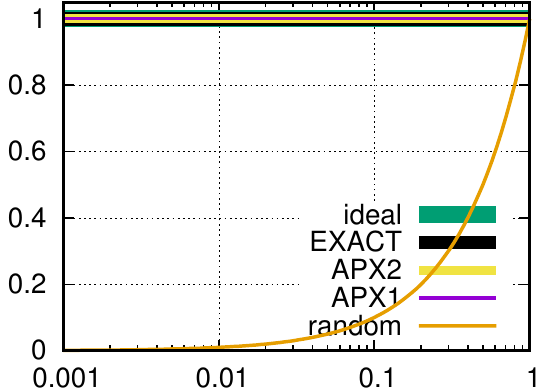}
   }\hfill
   \subfigure[Sign `zu']{%
      \includegraphics[scale=.77]{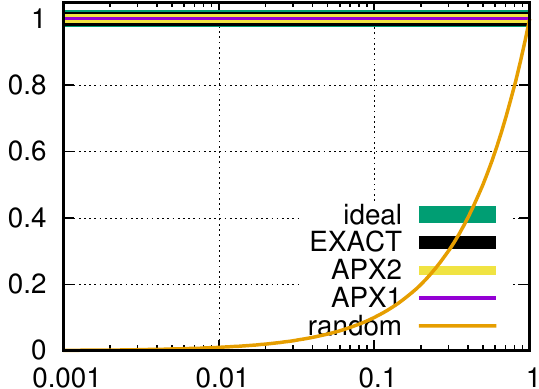}
   }
}
\end{figure}

\end{document}